\documentclass[11pt,a4paper]{article}
\usepackage{times,latexsym}
\usepackage{url}
\usepackage[T1]{fontenc}
\usepackage{graphicx}
\usepackage{tabularx}
\usepackage{amsmath}
\usepackage{amssymb}
\usepackage{booktabs}
\usepackage{algorithm}
\usepackage{algorithmic}
\usepackage{enumitem}
\usepackage{multirow}
\usepackage[acceptedWithA]{tacl2021v1}

\usepackage{xspace,mfirstuc,tabulary}

\newif\iftaclinstructions
\taclinstructionsfalse 
\iftaclinstructions
\renewcommand{\confidential}{}
\renewcommand{\anonsubtext}{(No author info supplied here, for consistency with
TACL-submission anonymization requirements)}
\newcommand{\instr}
\fi

\iftaclpubformat 

\else

\fi

\title{Prototype-Regularized Federated Learning for Cross-Domain Aspect Sentiment Triplet Extraction}

\author{
\begin{tabular}{c}
{\bfseries Zongming Cai$^{1}$ \qquad Jianhang Tang$^{1}$ \qquad Zhenyong Zhang$^{2}$} \\
{\bfseries Jinghui Qin$^{3}$ \qquad Kebing Jin$^{1}$\thanks{Corresponding author: kbjin@gzu.edu.cn} \qquad Hankz Hankui Zhuo$^{4}$} \\[0.4em]
$^{1}$State Key Laboratory of Public Big Data, Guizhou University \\
$^{2}$College of Computer Science and Technology, Guizhou University \\
$^{3}$School of Information Engineering, Guangdong University of Technology \\
$^{4}$School of Artificial Intelligence, Nanjing University
\end{tabular}
}

\date{}

\begin{document}
\maketitle

\begin{abstract}
Aspect Sentiment Triplet Extraction (ASTE) aims to extract all sentiment triplets of aspect terms, opinion terms, and sentiment polarities from a sentence. 
Existing methods are typically trained on individual datasets in isolation, failing to jointly capture the common feature representations shared across domains. Moreover, data privacy constraints prevent centralized data aggregation. To address these challenges, we propose Prototype-based Cross-Domain Span Prototype extraction (\textbf{PCD-SpanProto}), a prototype-regularized federated learning framework to enable distributed clients to exchange class-level prototypes instead of full model parameters. Specifically, we design a weighted performance-aware aggregation strategy and a contrastive regularization module to improve the global prototype under domain heterogeneity and the promotion between intra-class compactness and inter-class separability across clients. Extensive experiments on four ASTE datasets demonstrate that our method outperforms baselines and reduces communication costs, validating the effectiveness of prototype-based cross-domain knowledge transfer. 
\end{abstract}

\section{Introduction}
Aspect Sentiment Triplet Extraction (ASTE) is a subtask of aspect-based sentiment analysis (ABSA)~\cite{zhang2022survey}, aiming to extract sentiment triplets (i.e., aspect term, opinion term, sentiment polarity) from a sentence~\cite{peng2020knowing}. By jointly capturing what is being evaluated, how it is described, and the underlying sentiment, ASTE provides more opinion information than earlier subtasks such as aspect-level sentiment classification. Driven by the success of pre-trained language models and span-based architectures, recent methods have achieved remarkable success on standard benchmarks~\cite{liang2023stage,xu2021learning,nuo2024hybrid}. 

However, \textbf{existing models are trained and evaluated on individual datasets, thereby neglecting the latent common features and transferable knowledge shared across different domains.} In practice, different domains exhibit overlapping sentiment expression patterns. However, models optimized on a single domain are unable to adapt to unseen domains, as single-domain models suffer substantial performance deterioration when evaluated on out-of-domain data~\cite{xu2023measuring}.
An intuitive approach is to merge data across domains for joint training. However, raw data merging can introduce cross-domain noise that disrupts domain-specific pattern learning, making models that underperform single-domain-based approaches~\cite{wang2024refining}. Moreover, in realistic scenarios, centralized aggregation through data collection is often impractical due to privacy constraints and organizational boundaries.

To overcome those challenges of domain heterogeneity and data privacy, one way is to introduce a framework to share cross-domain knowledge, which neither merges raw data nor directly exposes private local data, e.g., Federated learning (FL)~\cite{mcmahan2017communication,rahman2025federated}.  It enables collaborative model training across distributed clients while preserving data privacy, where clients upload locally trained model parameters or locally computed gradients~\cite{hu2024practical,lu2025fedsmu} obtained from local training to the server to achieve aggregation and update the global model. However, conventional federated learning approaches that directly aggregate full model parameters face critical challenges in the context of ASTE. Firstly, transmitting the full set of model parameters per round incurs substantial communication cost, especially in the LLM-based natural language tasks. Secondly, the non-independent and identically distributed (non-IID) data across clients, where each domain exhibits distinct sentiment expressions client drift during local training, leading to aggregated models that converge slowly or fail to generalize across domains~\cite{lu2024federated}. Thirdly, conventional FL typically produces a unified global model shared by all clients, which is insufficient for capturing domain-specific characteristics and therefore cannot provide personalized adaptation for heterogeneous client domains. At last, directly averaging heterogeneous model parameters across domains may blur domain-specific representations rather than facilitate effective knowledge transfer~\cite{huang2023rethinking}.

Considering the non-IID challenge caused by heterogeneous domain distributions, personalized federated learning (pFL) has been introduced to preserve client-specific modeling ability while still benefiting from collaborative training~\cite{lu2024federated}. In particular, prototype-based federated learning has emerged as a promising solution, since it reduces communication overhead and alleviates heterogeneity through compact class-level representation sharing~\cite{tan2022fedproto,wu2024global,yin2025controlling}. However, existing pFL methods are largely designed for classification problems, where each instance is mapped to a single categorical label. This assumption does not hold for ASTE tasks, requiring structured span-level prediction over aspect terms, opinion terms, and their sentiment relations. Specifically, ASTE involves jointly identifying multiple spans, determining their boundaries, and modeling the relations among them to form sentiment triplets. As a result, key components in existing classification-oriented pFL frameworks, including prototype design, local training objectives, and global aggregation strategies, cannot be directly applied to ASTE. Although FL has recently been explored for several ABSA-related tasks, e.g., aspect category sentiment analysis~\cite{ahmad2024eafl} and document-level sentiment classification~\cite{ayman2026fedensemble}, these methods fundamentally focus on classification tasks and thus are inadequate for the structured extraction of ASTE.

Therefore, we propose \textbf{PCD-SpanProto} (\textbf{P}rototype-based \textbf{C}ross-\textbf{D}omain \textbf{Span} \textbf{Proto}type extraction), a prototype-regularized federated learning framework for cross-domain ASTE. Our framework builds on the STAGE span-level tagging model~\cite{liang2023stage} and introduces a prototype-based contrastive module that enables clients to exchange semantic knowledge across domains through prototype representations. Specifically, each client constructs local class-level prototypes from its span representations and uploads them to a central server, which aggregates them using performance-aware weights derived from clients' validation F1 scores. The aggregated global prototypes are then broadcast back to clients to regularize local training via contrastive alignment and separation losses, promoting intra-class compactness and inter-class separability across domains.

Our main contributions are as follows:
\begin{itemize}
    \item We identify the limitations of existing ASTE models that train on individual datasets in isolation, and propose a federated learning framework that enables cross-domain knowledge sharing without centralizing raw data.
    \item We introduce a prototype-based contrastive regularization strategy for the prediction of sentiment triplets, allowing clients to exchange class-level prototypes instead of full model parameters.
    \item Extensive experiments on four benchmark datasets demonstrate that our method consistently outperforms baselines, while the prototype-based communication component substantially reduces transmission cost.
\end{itemize}

\section{Related Work}
\subsection{Aspect-Based Sentiment Analysis (ABSA)}

Early approaches to Aspect Sentiment Triplet Extraction (ASTE) primarily adopted a pipeline framework. For example, Peng et al. ~\cite{peng2020knowing} proposed a two-stage method utilizing LSTMs to encode context and extract aspect and opinion terms independently, followed by a triplet pairing step. Subsequent studies improved feature representation by incorporating pre-trained language models (e.g., BERT~\cite{li2023double}and T5~\cite{gao2022lego}) to capture contextual semantics and Graph Neural Networks (GNNs)~\cite{chen2021semantic} to encode syntactic dependencies. However, these pipeline-based methods inevitably suffer from error propagation, thereby limiting overall performance.

\begin{figure*}[t]
    \centering
    \includegraphics[width=0.75\textwidth]{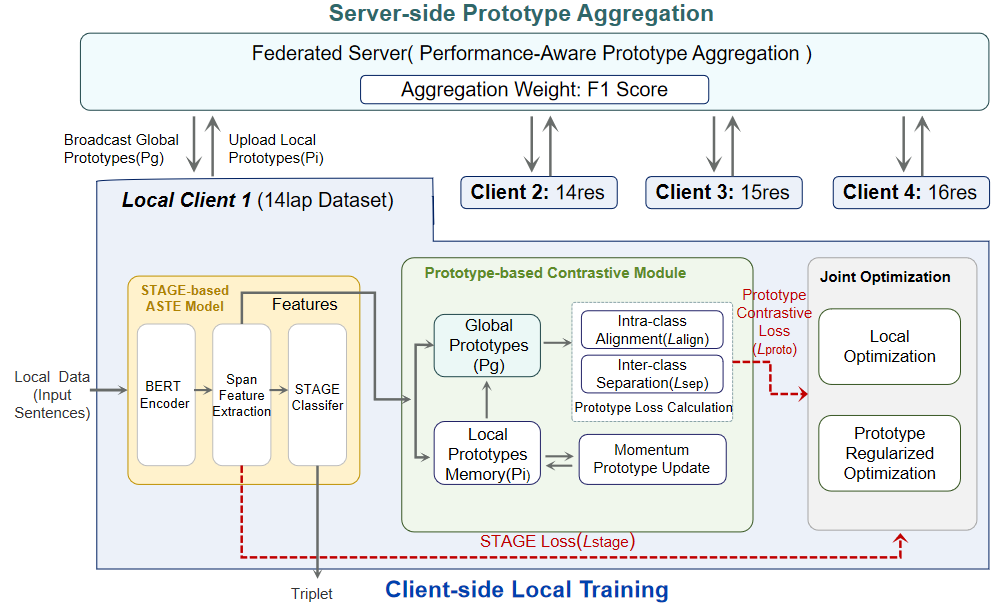}
    \vspace{-4mm}
    \caption{Overview of the Prototype-Regularized Federated Learning Framework for Cross-Domain ASTE.}
    \label{fig:framework}
    \vspace{-2mm}
\end{figure*}

To address this limitation, recent research has focused on end-to-end solutions. The table-filling paradigm~\cite{wu2020grid}, which formulates ASTE as a 2D grid tagging problem, has emerged as a prominent direction. Furthermore, acknowledging that aspect and opinion expressions often comprise multi-word phrases, researchers introduced span-based representation methods~\cite{chen2022span}~\cite{jin2024span}. By constructing span-level features rather than relying on single-token representations, they enable more precise modeling of entity boundaries~\cite{li2024improving}. To further capture complex dependencies between spans, recent works integrate GNNs~\cite{chen2022enhanced} or biaffine attention mechanisms ~\cite{nuo2024hybrid}. However, those models are trained on individual datasets, thereby neglecting the latent common features and transferable knowledge shared across different datasets. 

\subsection{Personalized Federated Learning (pFL)}
Federated learning (FL) ~\cite{mcmahan2017communication}enables collaborative model training across distributed clients while preserving data privacy. However, it suffers from severe statistical heterogeneity across different clients. Personalized federated learning (pFL) addresses this issue by learning client-specific models instead of a single global model. Existing pFL methods mainly include model mixing~\cite{hanzely2020federated}~\cite{deng2020adaptive}, regularization or multi-task learning~\cite{smith2017federated}~\cite{li2021ditto}, meta-learning~\cite{fallah2020personalized}, and parameter decoupling~\cite{arivazhagan2019federated}. These methods have achieved remarkable performance. However, many of them rely on heuristic designs or incur extra communication and computation costs. Recently, prototype-based pFL methods have explored sharing class-level representations rather than model parameters. FedProto~\cite{tan2022fedproto} aggregates class-wise feature prototypes to guide local training. However, most existing prototype-based methods~\cite{zhang2023efficient}~\cite{issaid2025tackling} are designed for standard classification tasks. They cannot be directly applied to structured prediction problems.

\section{The Proposed PCD-SpanProto}

The input of the model is a sentence $X = [x_1, x_2, \ldots, x_n]$, where $x_i$ denotes the $i$-th word in the sentence and $n$ is the sentence length. The objective of the Aspect Sentiment Triplet Extraction (ASTE) task is to extract $N$ sentiment triplets (aspect, opinion, sentiment), i.e., $T = \{(a_k, o_k, s_k)\}_{k=1}^{N}$, from $X$. The sentiment triplet consists of an aspect term $a_k$, its corresponding opinion term $o_k$, and a sentiment label $s_k \in \{\mathrm{POS}, \mathrm{NEU}, \mathrm{NEG}\}$ indicating the sentiment polarity expressed by the opinion toward the aspect, where $\mathrm{POS}$, $\mathrm{NEU}$, and $\mathrm{NEG}$ respectively denote positive, neutral, and negative sentiment.


As illustrated in Figure \ref{fig:framework}, we present \textbf{PCD-SpanProto}, a prototype-regularized federated learning framework for cross-domain aspect sentiment triplets extraction,  which consists of two components: a client-side local training module and a server-side prototype aggregation module.

\textbf{Client-side local training module} serves to extract information while preserving the privacy of client data, providing local knowledge to the global prototypes and cross-client knowledge integration. On each client, we adopt \textbf{a STAGE-based ASTE model} to encode local sentences and learn span-level representations, producing the primary task loss to guide the model to learn span-level representations. 
Additionally, we incorporate \textbf{a prototype-based contrastive module} that aligns local class-wise prototypes with global prototypes from the server. The generated prototype loss encourages consistency between local and global representations, promoting intra-class compactness and inter-class separability across clients. We jointly optimize them during local training, after which the updated local prototypes are uploaded to the server for global prototype aggregation.

\textbf{Server-side prototype aggregation module:}
This module integrates knowledge from all clients to guide local training. After each client completes local training, it uploads its class-level prototypes to the server. The server then aggregates these prototypes using F1-based weights derived from clients’ local validation performance, constructing global prototypes. Then the server broadcasts the aggregated global prototypes to all clients for the next round of federated optimization.

\subsection{Sentence Encoding and Span Representation}


At the beginning of local training, we encode the input sentences to obtain span-level features, used both for predicting local sentiment triplets and for constructing prototypes within the federated framework. To capture aspect terms, opinion terms, and sentiment relations, each input sentence $X = [x_1, x_2, \ldots, x_n]$ is first encoded using the BERT encoder to obtain contextualized token representations, then used to construct span-level representations. The encoded token representations are denoted as $H = [h_1, h_2, \ldots, h_m]$, where $h_i \in \mathbb{R}^{d}$ denotes the contextual embedding of the $i$-th token and $m$ represents the length of the tokenized sequence produced by BERT. 
However, the BERT tokenizer may split a single word into multiple subwords, leading to semantic integrity breaking and fragmented representations. Therefore, we reconstruct word-level representations by averaging the embeddings of its corresponding subwords to better capture the complete meaning of each word.
Specifically, if a word $x_i$ is tokenized into a set of subwords $\{h_k, h_{k+1}, \ldots, h_{k+t}\}$, its word-level representation $\tilde{h}_i$ is computed as:
\vspace{-1mm}
\begin{equation}
\tilde{h}_i = \frac{1}{t+1}\sum_{j=k}^{k+t} h_j 
\end{equation}

After this reconstruction step, we obtain the sequence of word-level representations:
$\tilde{H} = [\tilde{h}_1, \tilde{h}_2, \ldots, \tilde{h}_n].$
Then, we adopt a span-based representation strategy to capture multi-word aspects and opinion expressions. For a candidate span $s_{i,j}$ consisting of tokens from positions $i$ to $j$, we compute its representation using an attention-based aggregation mechanism. First, an attention score is assigned to each token within the span:
\vspace{-1mm}
\begin{equation}
\alpha_k = \frac{\exp(W_a \tilde{h}_k)}{\sum_{t=i}^{j} \exp(W_a \tilde{h}_t)}
\end{equation}
where $W_a$ is a trainable parameter. 
The span representation $s_{i,j}$ is obtained as a weighted combination of the token embeddings:
\vspace{-1mm}
\begin{equation}
s_{i,j} = \sum_{k=i}^{j} \alpha_k \tilde{h}_k 
\end{equation}

To improve computational efficiency, we apply a linear projection layer to map the span representation \(s_{i,j}\) into a lower-dimensional space. 
This results in the final span representation \(z_{i,j}\), as defined in Eq~(\ref{equation:span_representation}), where \(W_s\) and \(b_s\) are trainable parameters. And $z_{i,j}$ will be used for the prediction of sentiment triplets and the subsequent prototype construction and regularization process.
\vspace{-1mm}
\begin{equation}
z_{i,j} = W_s s_{i,j} + b_s
\label{equation:span_representation}
\end{equation}

\subsection{Prototype Construction and Contrastive Regularization}

Although federated learning provides an effective solution for enabling knowledge sharing across multiple datasets, directly sharing full model parameters is limited to the large model size and potential privacy risks~\cite{huang2023rethinking}. Therefore, we introduce a prototype-based representation learning mechanism that allows clients to exchange class-level prototypes instead of full model weights, reducing the communication overhead while preserving the cross-domain semantic information accuracy for effective ASTE.

Concretely, each prototype class corresponds to a tag $(t_a, t_o, t_s)$ indicating whether it is an aspect term, an opinion term, and the sentiment polarity. We assign each span representation $z_{i,j}$ obtained from Eq~(\ref{equation:span_representation}) to its corresponding prototype class based on the predicted tag. For example, $(N, O, \text{POS})$ denotes a span that serves as an opinion term in a positive aspect--opinion pair.

\paragraph{Span Tagging Scheme}

To define these spans, given an input sentence $X = [x_1, x_2, ..., x_n]$, we enumerate candidate spans $SP_{i,j}$ where $1 \le i \le j \le n$. Each $SP_{i,j}$ represents a contiguous segment of the sentence and serves as a potential aspect, opinion, or sentiment span.

Formally, the candidate span set is $SP = \{SP_{i,j} \mid 1 \le i \le j \le n \}$. Each span is then assigned a three-dimensional tag representing its roles in the ASTE task. Specifically, each span is labeled with a composite tag $(t_a, t_o, t_s)$ where $t_a \in \{A, N\}$, $t_o \in \{O, N\}$, and $t_s \in \{\text{POS}, \text{NEG}, \text{NEU}, N\}$. $t_a$ indicates whether the span is an aspect term. $t_o$ indicates whether the span is an opinion term. $t_s$ represents the sentiment polarity of the span. The meanings of the span-level tags are summarized in Table~\ref{tab:prototype_labels}.

\begin{table}[!ht]
\centering
\begin{tabularx}{\columnwidth}{cX}
\toprule
\textbf{Label} & \textbf{Meaning} \\
\midrule
A-O-S & span contains both an aspect and an opinion term, expressing a sentiment \\
N-O-S & span contains an opinion term, expressing a sentiment \\
A-N-S & span contains an aspect term, expressing a sentiment \\
N-N-S & span expresses a sentiment without containing aspect or opinion terms \\
N-N-N & span has no semantic role and carries no sentiment \\
\bottomrule
\end{tabularx}
\caption{Prototype labels derived from the three-dimensional span tagging scheme, where $t_s \neq N$ indicates the span expresses a sentiment.}
\label{tab:prototype_labels}
\end{table}

\begin{figure}[!t]
    \centering
    \includegraphics[scale=0.2]{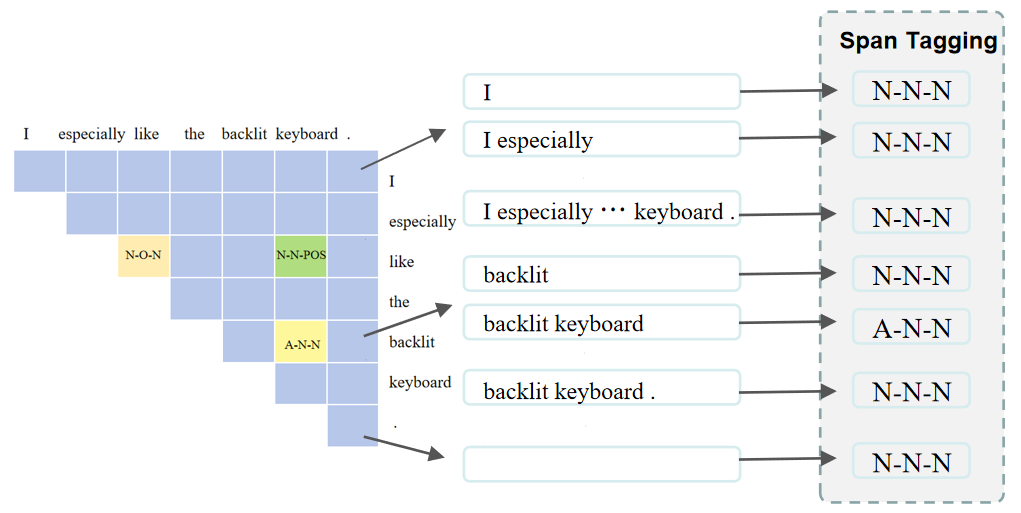}
    \caption{Illustration of the span tagging process.}
    \label{fig:span_tagging}
\end{figure}

Figure~\ref{fig:span_tagging} illustrates the span tagging process with a concrete example. \emph{For example, given a sentence ``I especially like the backlit keyboard.'', all candidate spans are enumerated and organized as an upper triangular span matrix. Each span is then classified into its corresponding tag. For instance, the span ``backlit keyboard'' is predicted as $A$-$N$-$N$, indicating it functions as an aspect term, while ``like the backlit keyboard'' is labeled as $N$-$N$-$POS$, representing a span expressing a positive sentiment toward the aspect.}

\paragraph{Local Prototype Construction}

Based on the span-level tags defined above, we construct local prototypes to summarize the representations of spans in each semantic category. Let $\mathcal{C}$ denote the set of all composite span tags $(t_a, t_o, t_s)$ from the span tagging scheme. Each $c \in \mathcal{C}$ corresponds to a specific aspect, opinion, or sentiment polarity. 

Given the span representations $z_{i,j}$ computed in Eq~\eqref{equation:span_representation}, we apply a classification layer to predict the probability of each tag for the span:
\vspace{-1mm}
\begin{equation}
p(y|z_{i,j}) = \text{Softmax}(W_z z_{i,j}+b)
\end{equation}
where $W_z$ denotes the classification parameters and $y$ represents one of the tag categories. For each category $c$, we collect all span representations which belong to that class and then compute the local prototype by averaging them:
\vspace{-1mm}
\begin{equation}
\hat{p}_c = \frac{1}{|S_c|} \sum_{z \in S_c} z
\label{eq:prototype_avg}
\end{equation}
where $S_c$ denotes the set of spans assigned to category $c$. 
After computing the local span prototypes $\hat{p}_c$, we update the prototype representations to enhance training robustness. However, directly using the current batch prototypes may lead to noisy or inaccurate prototype representations due to small batch sizes or feature variations. Therefore, we adopt a momentum-based update, which smooths the prototype over successive training steps, using the following rule:
\vspace{-1mm}
\begin{equation}
p_c^{(t)} = \beta p_c^{(t-1)} + (1-\beta)\hat{p}_c
\label{eq:prototype_momentum}
\end{equation}
where $\beta$ is the momentum coefficient and $t$ denotes the current training round.

\subsection{Performance-Aware Prototype Aggregation}

After each federated round, the server aggregates the local prototypes uploaded by all clients to update the global prototypes. A simple choice is to aggregate the prototypes with equal weights across clients. However, under domain heterogeneity, clients with weaker local performance may upload less reliable prototypes, and assigning them equal weight risks polluting the global prototype with noisy or poorly representations. To address this, the server collects the validation F1 scores from participating clients and adopts a performance-aware weighting strategy that assigns higher aggregation weights to clients with stronger local performance, reflecting their relative reliability. Let $F_k$ denote the validation F1 score of client $k$. The aggregation weight is defined as:

\vspace{-1mm}
\begin{equation}
w_k = \frac{F_k}{\sum_{i=1}^{K} F_i}
\label{eq:performance-aware
aggregation weight}
\end{equation}

Using these weights, the server aggregates the uploaded local prototypes to obtain a weighted global prototype for each category. Let $P_{k,c}$ denote the prototype of category $c$ uploaded from client $k$. The global prototype is computed as:
\vspace{-1mm}
\begin{equation}
P_c^{global} = \sum_{k=1}^{K} w_k P_{k,c}
\end{equation}
\label{eq:prototype_aggregation}
\subsection{Loss Computation}

To jointly optimize the span tagging task and prototype-based representation learning, we define a unified training objective that integrates the primary STAGE loss with a prototype-based contrastive regularization term. Building upon the span-level predictions and the constructed prototypes described above, the overall objective for each client is defined as the combination of the primary STAGE loss and the prototype-based contrastive regularization term:
\vspace{-1mm}
\begin{equation}
L = L_{\text{stage}} + \lambda L_{\text{proto}}
\label{eq:total_loss}
\end{equation}
where $\lambda$ is a balancing coefficient that scales the prototype regularization term.

\paragraph{Primary STAGE Loss}

Given span representations $z_{i,j}$, the model predicts the tag distribution via a softmax classifier and optimizes the corresponding cross-entropy loss, denoted as $L_{\text{stage}}$.
\begin{equation}
{\small L_{\text{stage}} = - \frac{1}{|SP|} \sum_{(i,j) \in SP} \sum_{c \in \mathcal{C}} y_{i,j,c} \log p(y{=}c \mid z_{i,j})}
\label{eq:stage_loss}
\end{equation}
where $SP$ denotes the set of all candidate spans in the sentence, 
$\mathcal{C}$ is the set of composite span tags, 
$y_{i,j,c}$ is the ground-truth indicator for class $c$, 
and $p(y=c \mid z_{i,j})$ represents the predicted probability of assigning tag $c$ to span $(i,j)$.
\paragraph{Prototype Contrastive Loss}
Let $P_c$ denote the global prototype of class $c\in\mathcal{C}$ broadcast by the server. For each span representation $z$ with label $y$, we encourage \emph{intra-class alignment} with $P_y$ and \emph{inter-class separation} from prototypes of other classes.
Concretely, we define:
\vspace{-1mm}
\begin{equation}
L_{\text{proto}} = \alpha L_{\text{align}} + \beta L_{\text{sep}}
\label{eq:proto_loss}
\end{equation}
where $\alpha$ and $\beta$ are balancing coefficients. The alignment loss encourages span representations to be close to the prototype of the category:
\vspace{-1mm}
\begin{equation}
L_{\text{align}} = -\cos(z, P_y)
\end{equation}
where $z$ denotes the span representation and $P_y$ represents the prototype of the ground-truth class.

The separation loss pushes the span representation away from prototypes of other classes:
\vspace{-1mm}
\begin{equation}
L_{\text{sep}} =
\log
\sum_{c\in\mathcal{C},\,c\neq y}
\exp(\cos(z,P_c))
\end{equation}

\subsection{Triplet Decoding}

Based on the predicted tags, we construct three span sets: the set of aspect spans $\mathcal{A} = \{SP_{i,j} \mid t_{a}^{i,j} = A\}$, the set of opinion spans $\mathcal{O} = \{SP_{i,j} \mid t_{o}^{i,j} = O\}$, and the set of sentiment spans $\mathcal{D} = \{(SP_{i,j}, s) \mid t_{s}^{i,j} = s,\ s \in \{\text{POS, NEG, NEU}\}\}$, where $i$ and $j$ denote the start and end token positions of a span, respectively.

For each sentiment span $(SP_{i,j}, s) \in \mathcal{D}$, we collect all candidate aspect spans $\mathcal{C}_A = \{SP_{i',j'} \in \mathcal{A} \mid i \leq i' \leq j' \leq j\}$ and opinion spans $\mathcal{C}_O = \{SP_{i',j'} \in \mathcal{O} \mid i \leq i' \leq j' \leq j\}$ whose boundaries fall within $[i, j]$. Here, $\mathcal{C}_A$ and $\mathcal{C}_O$ denote the sets of candidate aspect and opinion spans that are fully contained within the sentiment span $(i,j)$, where $(i',j')$ represents the start and end positions of a candidate span. Specifically, if both sets are non-empty, we form a triplet by selecting the aspect span with the largest right boundary $j'$ and the opinion span with the smallest left boundary $i'$, together with the sentiment polarity $s$. When the opinion span precedes the aspect span, the same procedure is applied symmetrically by exchanging the roles of $\mathcal{C}_A$ and $\mathcal{C}_O$.

\section{Experiments}
\subsection{Experimental Setup}
\paragraph{Datasets}
We conduct experiments on the ASTE-Data-V2 benchmark datasets~\cite{xu2020position}, consisting of one laptop-domain dataset (14Lap) and three restaurant-domain datasets (14Res, 15Res, and 16Res) derived from the SemEval Challenges. 
Since our method jointly trains on all four datasets on different clients, the original test instances may appear in the other training or validation datasets. To prevent data leakage, we remove all training or validation instances that overlap with the test datasets. Table~\ref{tab:dataset_stats} reports the dataset statistics before and after deduplication.

\begin{table}[!ht]
\centering
\resizebox{\columnwidth}{!}{
\begin{tabular}{l|ccc|ccc}
\toprule
\multirow{2}{*}{\textbf{Dataset}} & \multicolumn{3}{c|}{\textbf{Before}} & \multicolumn{3}{c}{\textbf{After}} \\
 & \textbf{Train} & \textbf{Val} & \textbf{Test} & \textbf{Train} & \textbf{Val} & \textbf{Test} \\
\midrule
14Lap & 906 & 219 & 328 & 906 & 219 & 328 \\
14Res & 1266 & 310 & 492 & 1265 & 310 & 492 \\
15Res & 605 & 148 & 322 & 605 & 147 & 322 \\
16Res & 857 & 210 & 326 & 599 & 145 & 326 \\
\bottomrule
\end{tabular}
}
\caption{Statistics before and after deduplication.}
\label{tab:dataset_stats}
\vspace{-1mm}
\end{table}

\paragraph{Implemention Details}
We implement our method across four datasets (14Lap, 14Res, 15Res, 16Res), each assigned to a distinct client. The STAGE model uses an uncased BERT-base encoder with a hidden dimension 768, producing logits for aspect, opinion, and sentiment spans. A prototype module is added with alignment loss weight 0.002, separation loss weight 0.00025, and EMA momentum 0.9. Training is conducted for 5 local epochs per client per federated round, with 50 rounds in total, and an AdamW optimizer with learning rates 2e-5 for BERT and 1e-3 for the classifier on a GPU (RTX 4090).

\paragraph{Baseline}

We evaluate the performance of PCD-SpanProto by comparing it against a set of baselines commonly used in ASTE and span-based extraction tasks, which can be broadly divided into two categories. The first category is pipeline methods, including Peng-two-stage \cite{peng2020knowing}, BMRC \cite{chen2021bidirectional}, and ASTE-RL \cite{yu2021aspect}, which extract aspects and opinions in separate stages. The second category is end-to-end methods, including GTS-BERT \cite{wu2020grid}, Span-ASTE \cite{xu2021learning}, EMC-GCN \cite{chen2022enhanced}, STAGE \cite{liang2023stage}, DRN \cite{xia2024dual}, ESSDB-GCN \cite{yang2024essdb}, MGEGCN \cite{tang2025multi}, MEGCN \cite{xu2025multiple}, S\&T \cite{nuo2024hybrid}, SE-TRDF \cite{liu2025sentiment}, and SimSTAR \cite{li2023simple}, which jointly predict sentiment triplets in a single framework.

\begin{table*}[t]
\centering
\resizebox{\textwidth}{!}{
\begin{tabular}{l|ccc|ccc|ccc|ccc}
\toprule
\textbf{Model} & \multicolumn{3}{c}{\textbf{14Lap}} & \multicolumn{3}{c}{\textbf{14Res}} & \multicolumn{3}{c}{\textbf{15Res}} & \multicolumn{3}{c}{\textbf{16Res}} \\
 & $P$ & $R$ & $F_1$ & $P$ & $R$ & $F_1$ & $P$ & $R$ & $F_1$ & $P$ & $R$ & $F_1$ \\
\midrule
Peng-two-stage$^\dagger$ & 37.38 & 50.38 & 42.87 & 43.24 & 63.66 & 51.46 & 48.07 & 57.51 & 52.32 & 46.96 & 54.21 & 50.41 \\
GTS-BERT$^\dagger$ & 58.62 & 52.35 & 55.29 & 68.14 & 68.77 & 68.45 & 62.37 & 59.71 & 60.98 & 66.16 & 68.81 & 67.44 \\
BMRC$^\dagger$ & 70.55 & 48.98 & 57.82 & 75.61 & 61.77 & 67.99 & 68.51 & 53.40 & 60.02 & 71.20 & 61.08 & 65.75 \\
ASTE-RL$^\dagger$ & 64.80 & 54.99 & 59.50 & 70.60 & 68.65 & 69.71 & 65.45 & 60.29 & 62.72 & 67.21 & 69.69 & 68.41 \\
Span-ASTE$^\dagger$ & 62.74 & 56.75 & 59.58 & 74.17 & 68.27 & 71.00 & 64.15 & 62.10 & 63.05 & 69.31 & 71.32 & 70.29 \\
EMC-GCN$^\dagger$ & 59.61 & 56.30 & 57.90 & 70.37 & 72.84 & 71.58 & 60.45 & 62.72 & 61.55 & 63.43 & 72.63 & 67.69 \\
STAGE$^\dagger$ & \underline{71.98} & 55.86 & 61.58 & 78.58 & 69.58 & 73.76 & \underline{73.63} & 57.90 & 64.79 & 76.67 & 70.12 & 73.24 \\
ESSDB-GCN$^\diamond$  & 71.76 & 55.13 & 62.38 & \textbf{79.69} & 68.93 & 73.99 & 72.57 & 57.69 & 64.36 & 74.43 & 67.21 & 70.63 \\
SimSTAR$^\diamond$  & 66.46 & \underline{58.23} & 62.07 & 76.23 & 71.63 & 73.86 & 71.71 & 59.59 & 65.09 & 72.02 & 74.12 & 73.06 \\
SE-TRDF$^\diamond$  & 66.89 & 48.89 & 56.49 & 73.03 & 65.84 & 69.24 & 69.47 & 54.43 & 61.04 & 73.02 & 64.78 & 68.65 \\
S\&T$^\diamond$  & 70.26 & 56.86 & 62.74 & 76.85 & 72.19 & 74.44 & 71.91 & 61.48 & 66.26 & \underline{76.72} & 70.50 & 73.47 \\
DRN$^\diamond$  & 66.99 & 52.61 & 58.94 & 75.24 & 64.49 & 69.45 & 68.61 & 55.47 & 61.34 & 73.30 & 64.42 & 68.57 \\
MEGCN$^\diamond$  & 68.83 & 57.26 & \underline{62.51} & 76.01 & 72.98 & 74.45 & 69.64 & \textbf{65.44} & 67.40 & 73.18 & 72.49 & 72.81 \\
MGEGCN$^\diamond$  & 66.04 & \textbf{58.60} & 62.10 & 76.14 & \underline{73.29} & \underline{74.69} & 71.49 & \underline{65.15} & \underline{68.18} & 72.74 & \underline{75.29} & \underline{74.00} \\
Ours & \textbf{72.32} & 55.64 & \textbf{63.66} & \underline{79.65} & \textbf{73.36} & \textbf{75.49} & \textbf{74.49} & 65.01 & \textbf{69.24} & \textbf{77.81} & \textbf{75.32} & \textbf{74.13} \\
$\Delta_\text{STAGE}$ & & & 2.08$\uparrow$ & & & 1.73$\uparrow$ & & & 4.45$\uparrow$ & & & 0.89$\uparrow$ \\
\bottomrule
\end{tabular}
}
\caption{Performance comparison of different models on ASTE datasets. $P$, $R$, and $F_1$ denote Precision, Recall, and F1-score. The second best F1 results are underlined. The symbol ``$\dagger$'' indicates results retrieved from \cite{liang2023stage}.The symbol ``$\diamond$ '' indicates results retrieved from their respective original papers.}
\label{tab:aste_results}
\end{table*}

\paragraph{Evaluation Protocol}
We evaluate triplet-level Precision, Recall, and F1-score as the evaluation metrics. In addition, we conduct four additional analyses to evaluate our framework:

 1. \textbf{Communication efficiency:} We analyze the reduction in transmitted parameters achieved by our prototype-based approach relative to conventional federated learning.

 2. \textbf{Training strategy:} We compare the federated 
learning against raw data merging and single-domain training to validate the advantage of cross-domain knowledge sharing.

 3. \textbf{Aggregation strategy:} We then compare our performance-aware F1-weighted aggregation against uniform aggregation to validate the adaptive client weighting under domain heterogeneity.

 4. \textbf{Knowledge transferring :} We evaluate the prototype representation effectiveness from cross-client prototype similarity, convergence behavior across federated rounds, and prototype representation quality via t-SNE visualization.

\subsection{Experimental Results}

Table~\ref{tab:aste_results} presents the performance of our proposed method compared with several baselines on the four ASTE datasets (14Lap, 14Res, 15Res, and 16Res), measured in terms of Precision, Recall, and F1-score.
Our method achieves the highest F1-scores across most datasets, demonstrating the effectiveness of incorporating prototype-based cross-domain knowledge into span-level ASTE. Compared with the STAGE baseline, our model produces consistent improvements across all four datasets, confirming that federated prototype regularization provides complementary supervision beyond what single-domain training can offer. 
Our approach further benefits from cross-domain prototype alignment, enabling more generalizable span representations without relying on explicit syntactic features.
Moreover, the balanced gains in both precision and recall suggest that prototype regularization improves not only the identification of valid triplet candidates but also the suppression of incorrect ones. These results suggest that the proposed framework can effectively leverage cross-domain prototypes and span-level representations to improve ASTE performance.

\subsection{Communication Efficiency of Prototype-Based Federated Learning}

Conventional federated learning requires transmitting the full model parameters at each round, which in our case amounts to over 110M parameters composing the BERT encoder and all classification layers. In contrast, our prototype-based approach requires each client to upload only a set of class-level prototypes, reducing the transmitted parameters to 3,200 per round. As shown in Table~\ref{tab:parameter_comparison}, the prototypes are comparable in size to the STAGE classifier alone (3,216 parameters), but carry semantic information from the different domains that classifier weights cannot provide. This drastic reduction in communication overhead makes our federated framework significantly more practical for real-world deployment, where bandwidth and privacy constraints are critical concerns, while still preserving the span-level representations for cross-domain ASTE.

\begin{table}[!ht]
\centering
\resizebox{\columnwidth}{!}{
\begin{tabular}{lc}
\toprule
\textbf{Type} & \textbf{Number of Parameters} \\
\midrule
Full STAGE Model & 110,298,760 \\
STAGE Classifier & 3,216 \\
Prototypes & 3,200 \\
\bottomrule
\end{tabular}
}
\caption{Comparison of parameter counts for different components and prototype-based communication.}
\label{tab:parameter_comparison}
\end{table}

\subsection{Cross-Domain Training Strategy Comparison}
To evaluate the benefits of cross-domain training, we conduct three experiments:

 1. \textbf{Merged single-model training:} All four datasets are combined, and a single Base STAGE model is trained on the merged data, 
    then evaluated on each dataset individually.
    
 2. \textbf{Single-domain base models:} Four independent Base STAGE 
    models are trained on each dataset separately. Each model is then 
    evaluated on all four datasets to assess its cross-domain generalization 
    capability.
    
 3. \textbf{Federated cross-domain training:} The four datasets are 
    treated as distinct clients. Each client trains a local model with 
    prototype-based regularization and is evaluated on all four datasets.

Table~\ref{tab:cross_domain} presents the triplet-level F1 scores across 
all three training strategies. The results show that: Firstly, \textbf{raw data merging degrades performance.} The Merged Single-Model 
underperforms the in-domain Base STAGE models across all 
datasets, confirming that directly combining heterogeneous datasets 
introduces cross-domain noise that disrupts domain-specific pattern learning. Secondly, \textbf{Single-domain models suffer from severe cross-domain degradation.} 
Each Base STAGE model performs well on its own domain but degrades drastically when evaluated on others, revealing that models trained in isolation learn domain-specific features that fail to transfer across domain boundaries. At last, \textbf{federated learning improves both in-domain and cross-domain performance.} Each client not only outperforms its Base STAGE counterpart on its own dataset, but also achieves substantial cross-domain gains compared to the corresponding single-domain model. This indicates that prototype-based FL enables effective cross-domain knowledge transfer, allowing each client to generalize beyond its local data distribution while maintaining in-domain specialization.
\begin{table}[!t]
\centering
\resizebox{\columnwidth}{!}{
\begin{tabular}{lcccc}
\toprule
Method & 14Lap & 14Res & 15Res & 16Res \\
\midrule
Merged Single-Model       & 61.42 & 72.05 & 61.35 & 72.23 \\
\midrule
Base STAGE (14Lap)        & \textbf{61.58} & 52.49 & 40.05 & 49.25 \\
Base STAGE (14Res)        & 40.10 & \textbf{73.76} & 58.58 & 64.62 \\
Base STAGE (15Res)        & 30.85 & 65.91 & \textbf{64.79} & 69.75 \\
Base STAGE (16Res)        & 32.98 & 65.26 & 62.85 & \textbf{73.24} \\
\midrule
Federated Client (14Lap)  & \textbf{63.66} & 54.33 & 56.96 & 66.24 \\
Federated Client (14Res)  & 42.66 & \textbf{75.49} & 59.45 & 66.26 \\
Federated Client (15Res)  & 40.80 & 67.34 & \textbf{69.24} & 72.29 \\
Federated Client (16Res)  & 42.86 & 67.13 & 68.67 & \textbf{74.13} \\
\bottomrule
\end{tabular}
}
\caption{Cross-domain evaluation results (triplet-level F1 scores). 
Bold values indicate in-domain performance for each model.}
\label{tab:cross_domain}
\vspace{-1mm}
\end{table}

\subsection{Effect of Performance-Aware Prototype Aggregation}

\begin{table}[!ht]
\centering
\resizebox{\columnwidth}{!}{
\begin{tabular}{lcccc}
\toprule
\textbf{Aggregation} & \textbf{14Lap} & \textbf{14Res} & \textbf{15Res} & \textbf{16Res} \\
\midrule
Uniform      & 61.89 & 75.12 & 68.71 & 73.84 \\
F1-weighted  & \textbf{63.66} & \textbf{75.49} & \textbf{69.24} & \textbf{74.13} \\
\bottomrule
\end{tabular}
}
\caption{Ablation study on prototype aggregation strategy (F1 scores).}
\label{tab:aggregation_ablation}
\end{table}

To validate the performance-aware aggregation strategy, we compare it against uniform aggregation, where the global prototype is computed as the simple average of all client prototypes, i.e., $w_k = \frac{1}{K}$ for all $k$. As shown in Table~\ref{tab:aggregation_ablation}, F1-weighted aggregation consistently outperforms uniform aggregation across all four datasets. The performance disparity is most evident on 14Lap, which exhibits the lowest cross-domain prototype similarity with the restaurant clients, as illustrated in Figure~\ref{fig:tsne_prototype}. Under uniform aggregation, prototypes from incompatible semantic subspaces are assigned equal influence over the global prototype, introducing noise that disrupts local representations. In contrast, the performance-aware weighting mechanism moderates the contribution of less compatible clients, preserving the integrity of cross-domain knowledge transfer.
These results confirm that adaptive weighting based on local validation performance is effective for prototype aggregation under domain heterogeneity.

\subsection{Effectiveness of Prototype-Based Cross-Domain Knowledge Transferring}

To further investigate how prototype-based federated learning facilitates cross-domain knowledge transfer, we analyze the problem from three perspectives: the pairwise similarity between client prototypes, the convergence behavior across federated rounds, and the geometric structure of prototype representations via t-SNE visualization. All prototype visualizations in the following experiments are based on the global prototypes obtained at the final federated round.

\subsubsection{Cross-Client Prototype Similarity}
\begin{figure}[!ht]
    \centering
    \includegraphics[width=0.65\columnwidth]{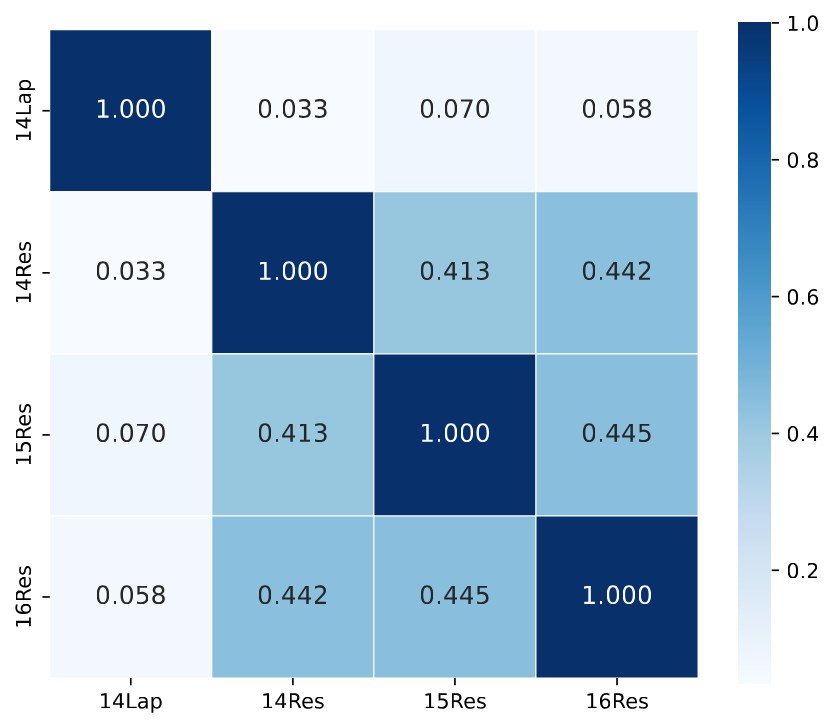}
    \caption{Cross-Domain prototype similarity among federated clients on four ASTE benchmarks}
    \label{fig:tsne_prototype}
\end{figure}
Figure~\ref{fig:tsne_prototype} presents the pairwise cosine similarity between the aggregated prototypes of all four clients. We observe a domain-driven similarity pattern.
This asymmetric distribution validates an assumption of our framework: intra-domain clients converge toward compatible prototype spaces, enabling more effective knowledge transfer through global aggregation, while inter-domain clients preserve domain-specific representations that maintain local specialization. 

\subsubsection{Convergence of Federated Training}

\begin{figure}[!ht]
    \centering
    \includegraphics[width=0.72\columnwidth]{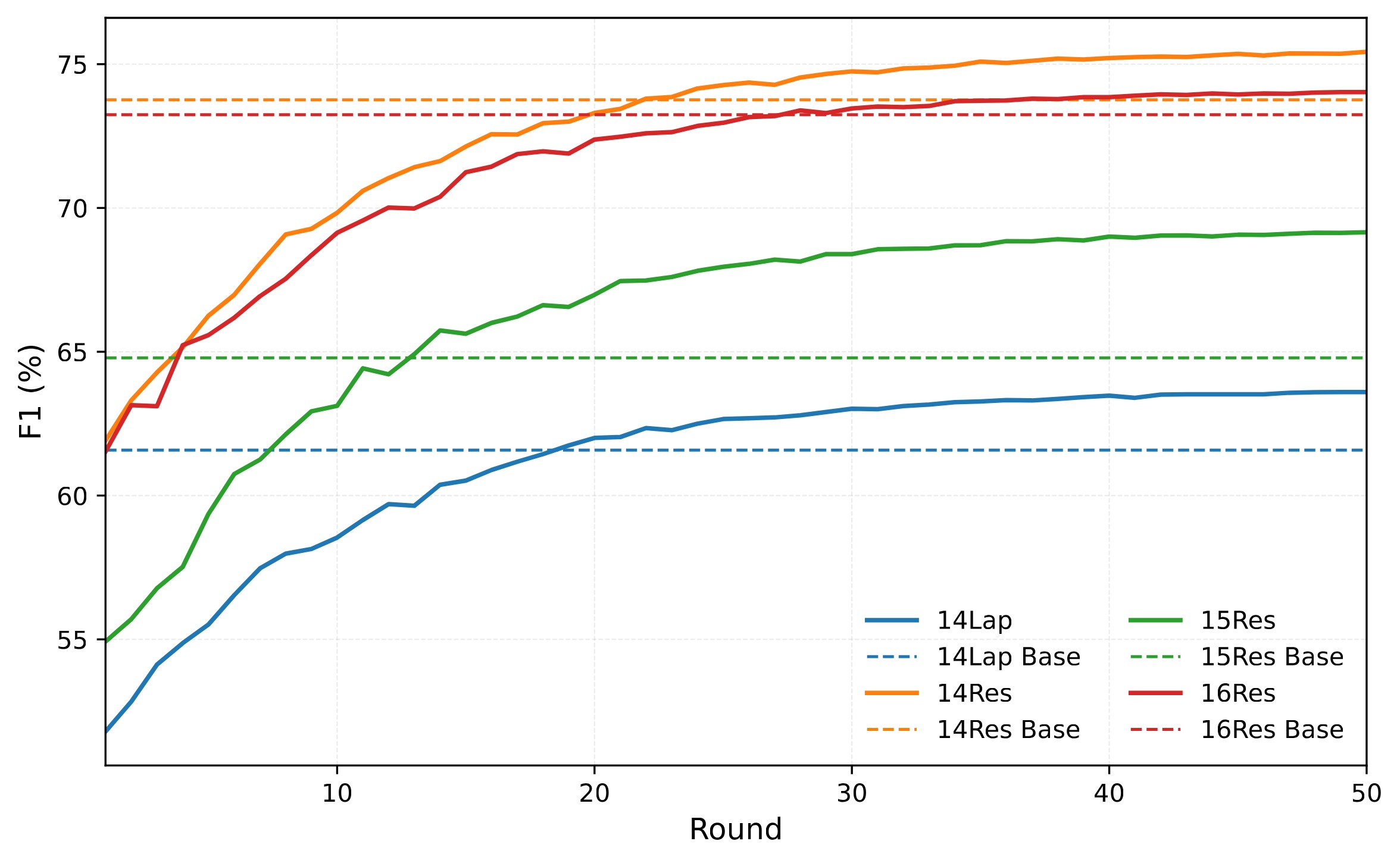}
    \caption{Triplet-level F1 scores of each client. }
    \label{fig:f1}
    \vspace{-3mm}
\end{figure}

Figure~\ref{fig:f1} illustrates the triplet-level F1 scores of each client across 50 federated rounds, with dashed lines indicating the corresponding Base STAGE final performance trained independently on each dataset. It is noted that clients in the restaurant domain converge more rapidly during the early training rounds, primarily due to the higher inter-client prototype similarity within the domain, thereby enabling the construction of more informative global prototypes from the local dataset. Furthermore, 14Lap exhibits a gradual improvement, consistent with the low cross-domain similarity illustrated in Figure~\ref{fig:tsne}. 
The smooth convergence of all clients further demonstrates the stabilizing effect of momentum-based prototype updates (Eq~\eqref{eq:prototype_momentum}) and performance-aware aggregation weights (Eq~\eqref{eq:performance-aware
aggregation weight}).
\subsubsection{Prototype Representation Quality}

\begin{figure}[!ht]
    \centering
    \includegraphics[width=0.75\columnwidth]{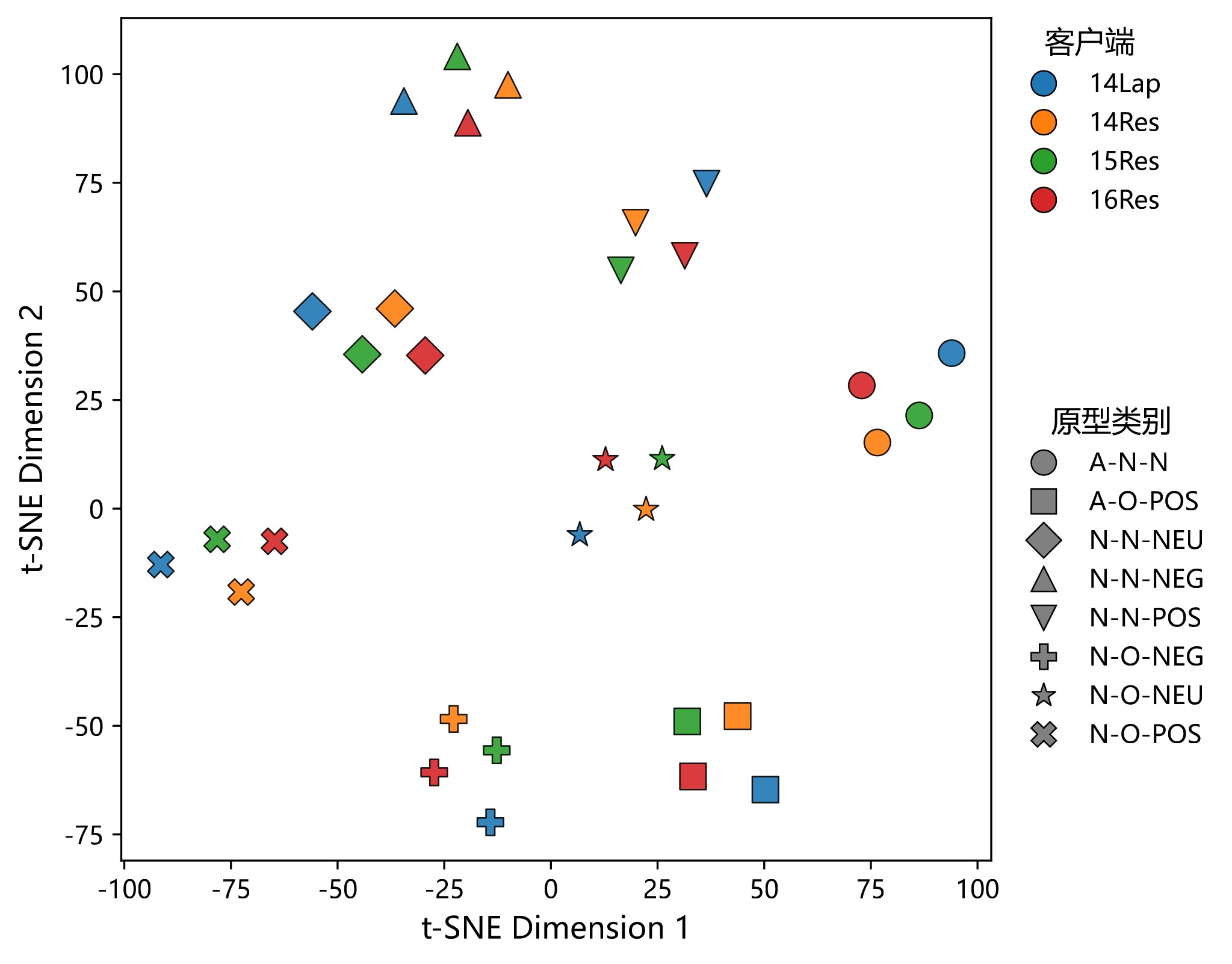}
    \caption{t-SNE visualization of 32 class-level prototype vectors 
(8 classes $\times$ 4 clients).  }
    \label{fig:tsne}
\end{figure}

To qualitatively assess the discriminability of learned prototype representations, we apply t-SNE to project the 32 class-level prototypes into two dimensions, as shown in Figure~\ref{fig:tsne}. 
The visualization reveals a clear class-driven clustering structure: prototypes of the same class tend to group together across clients, forming identifiable clusters in the projected space. 
However, the degree of clustering varies across classes. Semantically distinct classes such as N-N-NEU, N-N-NEG, and N-O-POS exhibit tighter cross-client groupings, suggesting that these categories carry more consistent semantic signals across domains. In contrast, the laptop-domain client (14Lap, blue) shows a deviation from the restaurant clients in several classes, which is consistent with the lower cross-domain prototype similarity observed in Figure~\ref{fig:tsne_prototype}. This confirms that while prototype-based federated learning promotes cross-client alignment, domain-specific variations are partially preserved, reflecting a balance between generalization and local specialization. Together, these observations provide qualitative support for the quantitative improvements reported in Table~\ref{tab:aste_results}.

\subsection{Parameter Sensitivity}

\begin{figure}[!ht]
    \centering
    \includegraphics[width=\columnwidth]{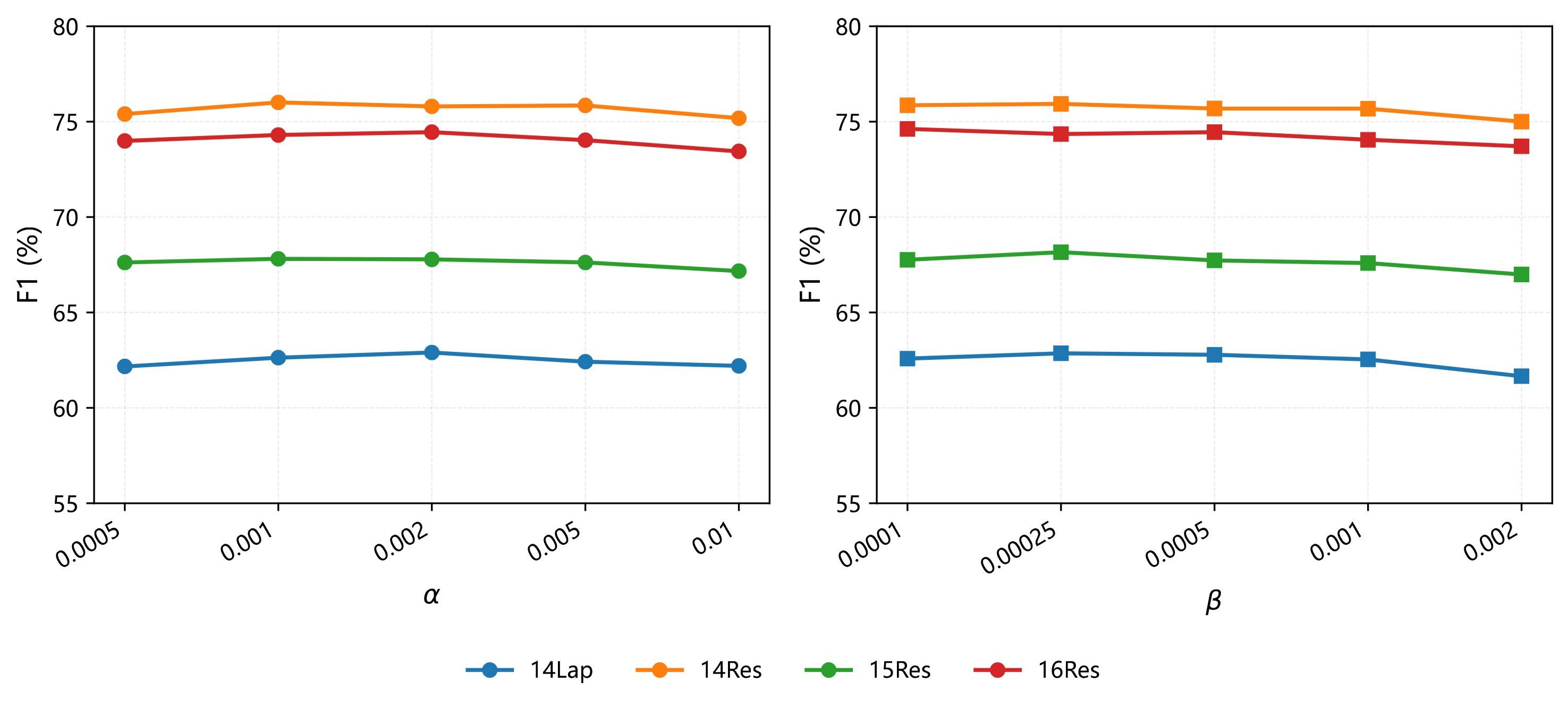}
    \vspace{-4mm}
    \caption{Parameter sensitivity analysis of alignment weight $\alpha$ 
    (left, $\beta$ fixed at $2.5\times10^{-4}$) and separation weight 
    $\beta$ (right, $\alpha$ fixed at $2\times10^{-3}$).}
    \label{fig:param}
\end{figure}

We investigate the sensitivity of the model to the prototype alignment weight $\alpha$ and separation weight $\beta$ in $\mathcal{L}_{\text{proto}}$ by varying one parameter while fixing the other at its optimal value, as shown in Figure~\ref{fig:param}.
In both cases, all four clients exhibit consistent trends across the search range, confirming that the optimal values provide a stable and effective configuration across heterogeneous domains.

\section{Conclusions}

In this paper, we propose PCD-SpanProto, a prototype-regularized federated learning framework for cross-domain Aspect Sentiment Triplet Extraction, enabling clients to upload prototypes rather than full model parameters. We introduce a prototype-based contrastive module to regularize local training through alignment and separation losses, and a performance-aware aggregation strategy to infuse global prototypes with cross-domain knowledge. Experiments demonstrate consistent improvements over single-domain baselines and competitive results against end-to-end methods, confirming that our method effectively transfers cross-domain knowledge without compromising data privacy or prohibitive communication costs. Future work will explore investigating more fine-grained prototype construction that better captures domain-specific linguistic variations within each semantic category.

\bibliography{tacl2021}
\bibliographystyle{acl_natbib}

\iftaclpubformat

\onecolumn

\fi

\end{document}